\documentclass[11pt]{article}
\usepackage{colacl}
\usepackage{isolatin1}
\usepackage{times}
\usepackage{latexsym}
\usepackage{epic}
\usepackage{ecltree}
\usepackage{eso-pic} 
\usepackage{graphicx} 
\AddToShipoutPicture{\put(40,750){\texttt{Third Int'l Workshop on Evaluating Word Sense Disambiguation Systems (SENSEVAL) 2004}}}

\title{WSD Based on Mutual Information and Syntactic Patterns}

\author{David Fernández-Amorós\\
  Departamento de Lenguajes y Sistemas Informáticos \\
  UNED\\
  {\tt david@lsi.uned.es}}

\date{}

\begin{document}

\maketitle

\begin{abstract}
  This paper describes a hybrid system for WSD, presented to the English all-words and lexical-sample tasks, that relies on two different unsupervised approaches. The first one selects the senses according to mutual information proximity between a context word a variant of the sense. The second heuristic analyzes the examples of use in the glosses of the senses so that simple syntactic patterns are inferred. This patterns are matched against the disambiguation contexts. We show that the first heuristic obtains a precision and recall of .58 and .35 respectively in the all words task while the second obtains .80 and .25. The high precision obtained recommends deeper research of the techniques.  Results for the lexical sample task are also provided.

\end{abstract}

\section{Introduction}

We will describe in this paper the system that we presented to the \textsc{SENSEVAL-3} competition in the English all-words and lexical-sample tasks. It is an unsupervised system that relies only on dictionary information and raw coocurrence data that we collected from a large untagged corpus. There is also a supervised extension of the system for the lexical sample task that takes into account the training data provided for the lexical sample task. We will describe two heuristics; the first one  selects the sense of the words' synset with a synonym with the highest Mutual Information (MI) with a context word.  This heuristic will be covered in section 2. The second heuristic relies on a set of syntactic structure rules that support particular senses. This rules have been extracted from the examples in WordNet sense glosses. Section 3 will be devoted to this technique. In section 4 we will explain the combination of both heuristics to finish in section 5 with our conclusions and some considerations for future work.

\section{Selection of the closest variant}

In the second edition of \textsc{SENSEVAL}, we presented a system, described in \cite{Fernandez-Amoros-01b}, that assigned scores to each word sense adding up Mutual Information estimates between all the pairs (word-in-context, word-in-gloss).  We have identified some problems with this technique.

\begin{itemize}

\item This exhaustive use of the mutual information estimates turned out to be very noisy, given that the errors in the individual mutual information estimates often correlated, thus affecting the final score for a sense. 

\item Sense glosses usually contain vocabulary that is not particularly relevant to the specific sense. 

\item Another typical problem for unsupervised systems is that the sense inventory contains many senses with little or no presence in actual texts. This last problem has been addressed in a very straightforward manner, since we have discarded the senses for a word with a relative frequency below 10\%.

\end{itemize}

The first problem might very well improve by itself when larger untagged corpora are available and increasing computing power  eliminates the need for a limited controlled vocabulary in the MI calculations. Anyway, a solution that we have tried to implement for this source of problems, that is, cumulative errors in estimates biasing the final result, consists in restricting the application of the MI measure to \emph{promising} candidates.

An interesting criterion for the selection of these candidates is to select those words in the context that form a collocation with the word to be disambiguated, in the sense that is defined in \cite{Yarowsky-93}. Yarowsky claimed that collocations are nearly monosemous, so identifying them would allow us to focus on very local context, which should make the disambiguation process, if not more efficient, at least easier to interpretate.

One  example of test item that was incorrectly disambiguated by the systems described in \cite{Fernandez-Amoros-01b}  is the word \emph{church} in the sentence~:

\emph{An ancient stone church stands amid the fields, the sound of bells cascading from its tower, calling the faithful to evensong.}

The applicable collocation here would be noun/noun so that \emph{stone} is the context word to be used. 

To address the second problem, the use of non-relevant words in the glosses, we have decided to consider only the variants (the synonyms in a synset,in the case of WordNet) of each sense.  These synonyms (i.e. variants of a sense) constitute the intimate matter of WordNet synsets, a change in a synset implies a change in the senses of the corresponding words, while the glosses are just additional information of secondary importance in the design of the sense inventory. To continue with the example,  the synonyms for the three synsets for church in WordNet are (excluding church itself, which is obviously common to all the synsets)~:
\begin{itemize}

\item Christian\_church $\rightarrow$ Christian (16), Christianity (11)
\item church\_building $\rightarrow$ building (187)
\item church\_service $\rightarrow$ service (6)

\end{itemize}

We didn't compute MI of compound words so instead we splitted them. Since church is the word to be disambiguated, \emph{Christian\_church} is converted to  \emph{church}, \emph{church\_building} to \emph{building} and \emph{church\_service} to \emph{service}. The numbers in parenthesis indicate the MI \footnote{for words $a$ and $b$, MI($a$,$b$)=$\frac{p(a \cap b)}{p(a) \cdot p(b)}$, the probabilities are estimated in a corpus.} between the term and \emph{stone}. In this case we have a clear and strong preference for the second sense, which happens to be in accordance with the gold standard.

Unfortunately, we didn't have the time to finish a collocation detection procedure, we just had enough time to POS-tag the text with the Brill tagger \cite{Brill-92}  and parse it with the Collins parser \cite{Collins-99}. That effort was put to use in the syntactic pattern-matching heuristic in the next section, so in this case we just limited ourselves to detect, for each variant, the context word with the highest MI.

It is important to note that this heuristic is not dependent on the glosses and it is completely unsupervised, so that it is possible to apply it to any language with a sense inventory based on variants, as is the case with the languages in EuroWordNet, and an untagged corpus.

We have evaluated this heuristic and the results are shown in table \ref{clovar_res}

\begin{table}[htbp]
{
\begin{center}
\begin{tabular}{|l|l|l|l|}
\hline
Task & Attempted & Prec & Recall\\ \hline
all words & 1215 / 2041 & .58 & .35\\ \hline
lexical sample & 938 / 3944 & .45 & .11\\ \hline
\end{tabular}
\caption{Closest variant heuristic results}
\label{clovar_res}
\end{center}
}
\end{table}

\section{Syntactic patterns}

\begin{figure}
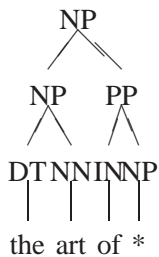

\begin{center}
\begin{bundle}{NP}
            \chunk{\begin{bundle}{NP}
                \chunk{\begin{bundle}{DT}
                    \chunk{the}
                    \end{bundle}}
                    \chunk{\begin{bundle}{NN}
                        \chunk{art}
                      \end{bundle}}
                \end{bundle}}
              \chunk{\begin{bundle}{PP}
                  \chunk{\begin{bundle}{IN}
                  \chunk{of}
                \end{bundle}} 
                 \chunk{\begin{bundle}{NP}
                     \chunk{*}
                   \end{bundle}}
                      \end{bundle}}
              \end{bundle}
              \caption{Example of syntactic pattern}
      \label{patternexample}
\end{center}              
\end{figure}

This heuristic exploits the regularity of syntactic patterns in sense disambiguation. These repetitive patterns effectively exist, although they might correspond to different word meanings . One example is the pattern in figure \ref{patternexample}

which usually corresponds to a specific sense of art in the \textsc{SENSEVAL-2} English lexical sample task.

This regularities can be attached to different degrees of specificity. One system that made use of these regularities is \cite{Tugwell-01}.  The regularities were determined by human interaction with the system. We have taken a different approach, so that the result is a fully automatic system. As in the previous heuristic, we didn't take into consideration the senses with a relative frequency below 10\%.

\begin{figure}
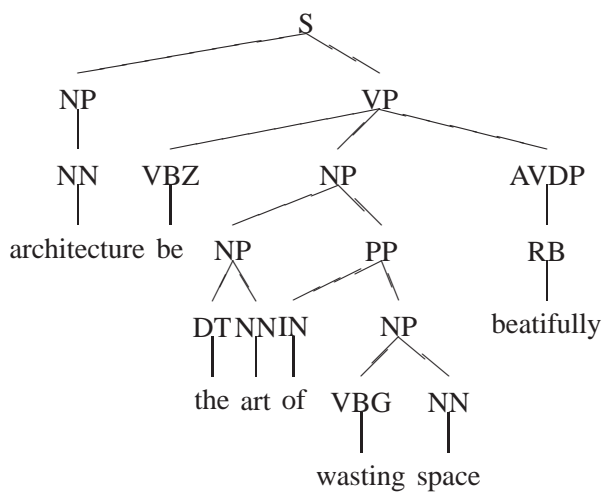

\begin{center}
  \begin{bundle}{S}
    \chunk{\begin{bundle}{NP} 
        \chunk{\begin{bundle}{NN}
        \chunk{architecture}
      \end{bundle}} 
  \end{bundle}} 
    \chunk{\begin{bundle}{VP}
        \chunk{\begin{bundle}{VBZ}
        \chunk{be}
      \end{bundle}} 
      \chunk{\begin{bundle}{NP}
            \chunk{\begin{bundle}{NP}
                \chunk{\begin{bundle}{DT}
                    \chunk{the}
                    \end{bundle}}
                    \chunk{\begin{bundle}{NN}
                        \chunk{art}
                      \end{bundle}}
                \end{bundle}}
              \chunk{\begin{bundle}{PP}
                  \chunk{\begin{bundle}{IN}
                  \chunk{of}
                \end{bundle}} 
                 \chunk{\begin{bundle}{NP}
                     \chunk{\begin{bundle}{VBG}
                     \chunk{wasting}
                     \end{bundle}}
                   \chunk{\begin{bundle}{NN}
                      \chunk{space}
                      \end{bundle}}
                      \end{bundle}}
                  \end{bundle}}
              \end{bundle}}
            \chunk{\begin{bundle}{AVDP}
                \chunk{\begin{bundle}{RB}
                \chunk{beatifully}
                \end{bundle}}
              \end{bundle}}
          \end{bundle}}
      \end{bundle}
      \caption{Top-level syntactic pattern}
      \label{primernivel}
\end{center}
\end{figure}

Due to time constraints we couldn't devise a method to identify salient syntactic patterns useful for WSD, although the task seems challenging. Instead, we parsed the examples in WordNet glosses. These examples are usually just phrases, not complete sentences, but they can be used as patterns straightaway. We parsed the test instances as well and looked for matches of the example inside the parse tree of the test instance. Coverage was very low. In order to increase it, we adopted the following strategy~: To take a gloss example and go down the parse tree looking for the word to disambiguate. The subtrees of the visited nodes are smaller and smaller. Matching the whole syntactic tree of the example  is rather unusual but chances increase with each of the subtrees. Of course, if we go too low in the tree we will be left with the single target word, which should in principle match all the corresponding trees of the test items of the same word. We will illustrate the idea with an example. An example of an \emph{art} sense gloss is : \emph{Architecture is the art of wasting space beatifully}.  We can see the parse tree depicted in figure  \ref{primernivel}.

\begin{figure}
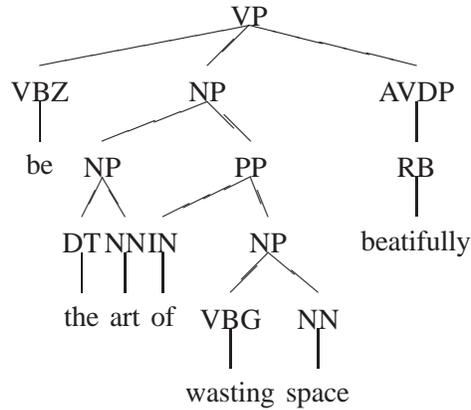

\begin{center}
\begin{bundle}{VP}
        \chunk{\begin{bundle}{VBZ}
        \chunk{be}
      \end{bundle}} 
      \chunk{\begin{bundle}{NP}
            \chunk{\begin{bundle}{NP}
                \chunk{\begin{bundle}{DT}
                    \chunk{the}
                    \end{bundle}}
                    \chunk{\begin{bundle}{NN}
                        \chunk{art}
                      \end{bundle}}
                \end{bundle}}
              \chunk{\begin{bundle}{PP}
                  \chunk{\begin{bundle}{IN}
                  \chunk{of}
                \end{bundle}} 
                 \chunk{\begin{bundle}{NP}
                     \chunk{\begin{bundle}{VBG}
                     \chunk{wasting}
                     \end{bundle}}
                   \chunk{\begin{bundle}{NN}
                      \chunk{space}
                      \end{bundle}}
                      \end{bundle}}
                  \end{bundle}}
              \end{bundle}}
            \chunk{\begin{bundle}{AVDP}
                \chunk{\begin{bundle}{RB}
                \chunk{beatifully}
                \end{bundle}}
              \end{bundle}}
          \end{bundle}
          \caption{Second syntactic pattern}
          \label{segundonivel}
\end{center}
\end{figure}

We could descend from the root, looking for the occurrence of the target word and obtain a second, simpler, pattern, shown in figure \ref{segundonivel}.

\begin{figure}
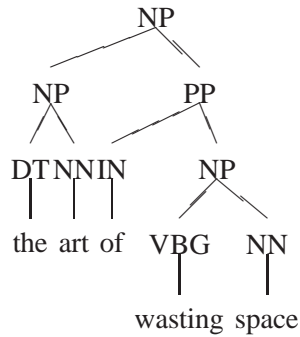

\begin{center}
\begin{bundle}{NP}
            \chunk{\begin{bundle}{NP}
                \chunk{\begin{bundle}{DT}
                    \chunk{the}
                    \end{bundle}}
                    \chunk{\begin{bundle}{NN}
                        \chunk{art}
                      \end{bundle}}
                \end{bundle}}
              \chunk{\begin{bundle}{PP}
                  \chunk{\begin{bundle}{IN}
                  \chunk{of}
                \end{bundle}} 
                 \chunk{\begin{bundle}{NP}
                     \chunk{\begin{bundle}{VBG}
                     \chunk{wasting}
                     \end{bundle}}
                   \chunk{\begin{bundle}{NN}
                      \chunk{space}
                      \end{bundle}}
                      \end{bundle}}
                  \end{bundle}}
              \end{bundle}
              \caption{third syntactic pattern}
      \label{tercernivel}
\end{center}              
\end{figure}

\begin{figure}
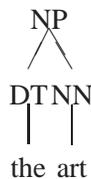

\begin{center}
\begin{bundle}{NP}
                \chunk{\begin{bundle}{DT}
                    \chunk{the}
                    \end{bundle}}
                    \chunk{\begin{bundle}{NN}
                        \chunk{art}
                      \end{bundle}}
                \end{bundle}
                \caption{fourth syntactic pattern}
                \label{cuartonivel}
\end{center}
\end{figure}

\begin{figure}
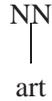

\begin{center}
  \begin{bundle}{NN}
    \chunk{art}
  \end{bundle}
  \caption{fifth syntactic pattern}
  \label{quintonivel}
\end{center}
\end{figure}

Following the same procedure we would acquire the patterns shown in figures \ref{tercernivel} y \ref{cuartonivel}, and the we would be left with mostly useless pattern shown in figure \ref{quintonivel}

Since there is an obvious tradeoff between coverage and precision, we have only made disambiguation rules based on the first three syntactic levels, and rejected rules with a pattern with only one word.

Still, coverage seems to be rather low and there are areas of the pattern that look like they could be generalized without much loss of precision, even when it might be difficult to identify them. Our hypothesis is that function words play an important role in the discovery of these syntactic patterns. We had no time to further investigate the fine-tuning of these patterns, so we added a series of transformations for the rules already obtained. In the first place, we replaced every tagged pronoun form with a wildcard meaning that every word tagged as a pronoun would match. In order to increase even more the number of rules we derive more rules keeping the part-of-speech tags and replacing content words with wilcards. 

We wanted to derive a larger set of rules, with the two-fold intention of achieving increased coverage and also to test if the approach was feasible with a rule set in the order of the hundreds of thousands or even millions. Every rule specifies the word for which it is applicable (for the sake of efficiency) and the sense the rule supports, as well as the syntactic pattern. We derived new rules in which we substituted the word to be disambiguated for each of its variants in the corresponding sense (i.e. the synonyms in the corresponding synset). The substitution was carried out sensibly in all the four fields of the rule, with the new word-sense (corresponding to the same synset as the old one), the new variant and the new syntactic pattern. This way we were able to effectively multiply the size of the rule set.

We have also derived a set of disambiguation rules based on the training examples for the English lexical sample task. The final rule  set consists of more than 300000 rules. The score for a sense is determined by the total number of rules it matches. We only take the sense with the highest score.

The results of the evaluation for this heuristic are shown in table \ref{synpat_res}

\begin{table}[htbp]
{
\begin{center}
\begin{tabular}{|l|l|l|l|}
\hline
Task & Attempted & Prec & Recall\\ \hline
all words & 648 / 2041 & .80 & .25\\ \hline
lexical sample & 821 / 3944 & .51 & .11\\ \hline
\end{tabular}
\caption{Syntactic pattern heuristic results}
\label{synpat_res}
\end{center}
}
\end{table}

\section{Combination}

Since we are interested in achieving a high recall and both our heuristics have low coverage, we decided to combine the results in a blind way with the first sense heuristic. We did a linear combination of the three heuristics, weighting the three of them equally, and returned the sense with the highest score. 

\section{Conclusions and future work}

The official results clearly show that the dependency of the system on  the first sense heuristic is very strong. We should have been more confident in our heuristics so that maybe a linear combination giving more weight to them in opposition to the first sense baseline would have produced better results.  The supervised extension of the algorithm, in which the syntactic patterns are learnt from the training examples as well as from the synset's glosses doesn't offer any improvement at all. The simple explanation is that the increase in the number of rules from the unsupervised heuristic to the supervised extension is only 17\% so no changes are noticeable at the answer level.

The results for the two heuristics are very encouraging. There are several points that deserve further investigation. It should be relatively easy to detect Yarowsky's collocations from a parse tree and that is likely to offer even better results in terms of precision, although the potential for increased coverage is unclear. As far as the other heuristic is concerned, it seems worthwhile to spend some time determining syntactic patterns more accurately. A good point to start could be statistical language modeling over large corpora, now that we have adapted the existing resources and parsing massive text collections is relatively easy. Of course, a WSD system aimed for final applications should also take advantage of other knowledge sources researched in previous work.

\bibliographystyle{acl}
\bibliography{my-citations}     

\begin{thebibliography}{}

\bibitem[\protect\citename{Brill}1992]{Brill-92}
Eric Brill.
\newblock 1992.
\newblock {A simple rule-based part-of-speech tagger}.
\newblock In {\em Proceedings of {ANLP}-92, 3rd Conference on Applied Natural
  Language Processing}, pages 152--155, Trento, IT.

\bibitem[\protect\citename{Collins}1999]{Collins-99}
Michael Collins.
\newblock 1999.
\newblock {\em Head-Driven Statistical Models for Natural Language Parsing}.
\newblock {Ph.D.} thesis, University of Pennsylvania.

\bibitem[\protect\citename{Fernández-Amorós \bgroup et al.\egroup
  }2001]{Fernandez-Amoros-01b}
D.~Fernández-Amorós, J.~Gonzalo, and F.~Verdejo.
\newblock 2001.
\newblock {The UNED systems at SENSEVAL-2}.
\newblock In {\em Second International Workshop on Evaluating Word Sense
  Disambiguation Systems (SENSEVAL), Toulouse}, pages 75--78.

\bibitem[\protect\citename{Tugwell and Kilgarriff}2001]{Tugwell-01}
David Tugwell and Adam Kilgarriff.
\newblock 2001.
\newblock Wasp-bench : A lexicographic tool supporting word sense
  disambiguation.
\newblock In David Yarowsky and Judita Preiss, editors, {\em {Second
  International Workshop on Evaluating Word Sense Disambiguation Systems
  (SENSEVAL), Toulouse}}, pages 151--154.

\bibitem[\protect\citename{Yarowsky}1993]{Yarowsky-93}
David Yarowsky.
\newblock 1993.
\newblock {One Sense per Collocation}.
\newblock In {\em {Proceedings, ARPA Human Language Technology Workshop}},
  pages 266--271, Princeton.

\end{thebibliography}

\end{document}